\pdfoutput=1
\documentclass[11pt]{article}
\usepackage{graphicx}

\usepackage[final]{acl}

\usepackage{times}
\usepackage{latexsym}
\usepackage{pdfpages}
\usepackage{tabularx}
\usepackage{booktabs}
\usepackage{array}
\usepackage{multirow}
\usepackage{ulem}
\usepackage{pifont}
\usepackage[section]{placeins}
\usepackage{balance}

\usepackage[T1]{fontenc}
\usepackage[utf8]{inputenc}

\usepackage{microtype}

\usepackage{inconsolata}

\newcommand{\cmark}{\ding{51}}%
\newcommand{\xmark}{\ding{55}}%

\title{Dense Passage Retrieval:\\Is it 
Retrieving?}

\author{Benjamin Reichman \\
  Georgia Institute of Technology \\
  \texttt{bzr@gatech.edu} \\\And
  Larry Heck \\
  Georgia Institute of Technology  \\
  \texttt{larryheck@gatech.edu} \\}

\begin{document}
\maketitle
\begin{abstract}
Large Language Models (LLMs) internally store repositories of knowledge. However, their access to this repository is imprecise and they frequently hallucinate information that is not true or does not exist. A paradigm called Retrieval Augmented Generation (RAG) promises to fix these issues. Dense passage retrieval (DPR) is the first step in this paradigm. In this paper, we analyze the role of DPR fine-tuning and how it affects the model being trained. DPR fine-tunes pre-trained networks to enhance the alignment of the embeddings between queries and relevant textual data. We explore DPR-trained models mechanistically by using a combination of probing, layer activation analysis, and model editing. Our experiments show that DPR training \textbf{decentralizes} how knowledge is stored in the network, creating \textbf{multiple access pathways} to the same information. We also uncover a \textbf{limitation} in this training style: the \textbf{internal knowledge} of the pre-trained model \textbf{bounds} what the retrieval model can retrieve. These findings suggest a few possible directions for dense retrieval: (1) expose the DPR training process to more knowledge so more can be decentralized, (2) inject facts as decentralized representations, (3) model and incorporate knowledge uncertainty in the retrieval process, and (4) directly map internal model knowledge to a knowledge base.  
\end{abstract}

\section{Introduction}
In just a few years, Large Language Models (LLMs) have emerged from research labs to become a tool utilized daily by hundreds of millions of people and integrated into a wide variety of businesses. Despite their popularity, these models have been critiqued for frequently hallucinating, confidently outputting incorrect information \cite{chatgpt_hallucinations}. Such inaccuracies not only mislead people but also erode trust in LLMs. Trust in these systems is crucial to their success and rate of adoption.

The retrieval augmented generation (RAG) paradigm is an approach to address  hallucinations~\cite{rag}. Unlike traditional LLM interactions where a query directly prompts an output from the model, an intermediary step is introduced. Initially, a "retrieval" model processes the query to gather additional information from a knowledge base, such as Wikipedia or the broader internet. This additional information alongside the original query is fed to the LLM, increasing the accuracy of the answers that the LLM generates. 

For this paradigm to be effective, the underlying retrieval model has to excel at finding accurate and relevant information. Typically, model performance is evaluated based on metrics that consider the top-5, top-20, top-50, and top-100 retrieved passages. However, recent studies indicate that LLMs predominantly use information from the top-1 to top-5 passages, underscoring the importance in RAG of not only high recall in retrieval but also precision in ranking \cite{lostinmiddle, xu2024retrieval}. 
%Retrieval model performance improves greatly as one goes from top-1 to top-100, highlighting a central issue in the current RAG pipeline. 
One approach to achieve both high recall and precision involves integrating a "reranking" model, which adjusts the order of retrieved passages to improve the relevance of the top-ranked passages \cite{reranking1, reranking2}. However, this approach adds the computational and maintenance cost of an additional model to the pipeline and can also introduce errors. The alternative option is to improve retrieval models to directly retrieve and rank passages well.

\begin{table*}[htbp]
\centering
{
\small
\begin{tabular}{m{1.1cm}|m{2.3cm}m{0.45cm}m{0.45cm}m{0.45cm}m{0.45cm}m{0.45cm}m{0.45cm}m{0.45cm}m{0.45cm}m{0.45cm}m{0.45cm}m{0.45cm}m{0.45cm}m{0.45cm}}
\toprule
Task & Model & Layer 0 & Layer 1 & Layer 2 & Layer 3 & Layer 4 & Layer 5 & Layer 6 & Layer 7 & Layer 8 & Layer 9 & Layer 10 & Layer 11 & Layer 12 \\
\hline
\multirow{4}{1.5cm}[-1.5ex]{2-Passage Probing}
& Pre-trained BERT – Untrained Probe & 0.50 & 0.50 & 0.51 & 0.48 & 0.50 & 0.52 & 0.51 & 0.51 & 0.50 & 0.49 & 0.50 & 0.54 & 0.50 \\
& Pre-trained BERT & 0.51 & 0.69 & 0.74 & 0.74 & 0.77 & 0.79 & 0.81 & 0.81 & 0.81 & 0.82 & 0.83 & 0.84 & 0.84  \\
& DPR-BERT Query Model & 0.51 & 0.68 & 0.74 & 0.77 & 0.79 & 0.80 & 0.81 & 0.83 & 0.82 & 0.83 & 0.83 & 0.82 & 0.82 \\
& DPR-BERT Context Model & 0.51 & 0.68 & 0.74 & 0.77 & 0.79 & 0.80 & 0.81 & 0.83 & 0.82 & 0.83 & 0.83 & 0.82 & 0.82 \\
\hline
\multirow{2}{1.5cm}{3-Passage Probing} & 
%Pre-trained BERT – Untrained Probe & 0.33 & 0.33 & 0.34 & 0.34 & 0.33 & 0.34 & 0.36 & 0.34 & 0.36 & 0.32 & 0.35 & 0.35 & 0.34 \\
Pre-trained BERT & 0.34 & 0.53 & 0.59 & 0.59 & 0.65 & 0.64 & 0.67 & 0.67 & 0.68 & 0.69 & 0.69 & 0.73 & 0.73 \\
& DPR-BERT & 0.34 & 0.54 & 0.60 & 0.63 & 0.66 & 0.66 & 0.66 & 0.70 & 0.71 & 0.69 & 0.73 & 0.72 & 0.71 \\
\hline
\multirow{2}{1.5cm}{4-Passage Probing}
% Pre-trained BERT – Untrained Probe & 0.25 & 0.25 & 0.24 & 0.25 & 0.26 & 0.25 & 0.25 & 0.26 & 0.25 & 0.25 & 0.24 & 0.25 & 0.24 \\
& Pre-trained BERT & 0.26 & 0.43 & 0.47 & 0.49 & 0.53 & 0.57 & 0.61 & 0.60 & 0.56 & 0.62 & 0.64 & 0.66 & 0.66 \\
& DPR-BERT & 0.26 & 0.46 & 0.51 & 0.54 & 0.57 & 0.58 & 0.60 & 0.63 & 0.64 & 0.63 & 0.65 & 0.63 & 0.63 \\
\hline
\multirow{2}{1.5cm}{5-Passage Probing} 
% Pre-trained BERT – Untrained Probe & 0.20 & 0.21 & 0.20 & 0.20 & 0.20 & 0.20 & 0.20 & 0.20 & 0.20 & 0.21 & 0.21 & 0.20 & 0.21 \\
& Pre-trained BERT & 0.21 & 0.35 & 0.42 & 0.43 & 0.43 & 0.50 & 0.53 & 0.53 & 0.54 & 0.56 & 0.57 & 0.60 & 0.61 \\
& DPR-BERT & 0.21 & 0.36 & 0.42 & 0.48 & 0.49 & 0.51 & 0.54 & 0.56 & 0.58 & 0.58 & 0.60 & 0.56 & 0.56 \\
\bottomrule
\end{tabular}
}
\caption{This table presents the outcomes of linear probing, where probes classify 2 to 5 passages to determine the best match for a given query. Due to identical performance metrics, DPR-BERT Query and Context model results are consolidated and displayed only for the 2-Passage Probe. Given that probes without training achieved performance at random chance levels across all passage counts, their results are reported solely for the 2-Passage Probe for comparison.}
\label{tab:probing}
\end{table*}

Retrieval methods can be broadly categorized into two types: sparse and dense \cite{retrievalsurvey}. Sparse methods encode queries and passages into sparse vectors, usually based on terms that appear in the queries and passages \cite{bm25,tfidf}. Dense methods employ language models to encode the semantic information in queries and passages into dense vectors \cite{dpr,dssm2013}. Dense methods often share two common properties: (a) the joint training of two or more encoding models – one for embedding a query and the other for embedding a knowledge base, and (b) contrastive training. These commonalities were introduced in the DPR method, inspiring many subsequent methods in the literature.

In this paper, we analyze the original DPR method using the BERT-base backbone. Through the experiments in each section, we analyze DPR from multiple perspectives to understand what is changing in the backbone model during the training process. We find that:
\begin{enumerate}
    \item BERT's ability to discriminate between passages remains unchanged (Section \ref{sec:probing}).
    \item The way that knowledge is structured in BERT changes to be more decentralized after DPR training (Section \ref{sec:knowledge_decentralization}).
    \item The knowledge that BERT can retrieve over is an extension of the knowledge contained within it (Section \ref{sec:model_editing}).
\end{enumerate}

%We begin by probing the model to determine if the features of pre-trained BERT are as discriminative as DPR-BERT in matching a query to the correct passage amongst hard-negative passages (Section \ref{sec:probing}). Next, using techniques from the pruning literature, we compare the relative strength and number of activations of the feedforward layers throughout the original pre-trained and DPR-trained models (Section~\ref{sec:knowledge_decentralization}). Finally, we add and remove knowledge from the network to investigate how knowledge interacts with DPR training (Section \ref{sec:model_editing}). 

\section{Knowledge Consistency Between Untrained and Trained Model}
\label{sec:probing}
Language models are known to store a vast amount of knowledge, with the feedforward layers of the transformer architecture acting as a key-value memory store of knowledge \cite{ffl_kvm}. This section details experiments conducted to understand the impact of DPR-style training from a model-knowledge perspective.

Linear probing, a method to characterize model features, involves training a linear classifier on the internal activations of a frozen network to execute a simple task \cite{bengioprobing}. This reveals the mutual information shared between the model's primary training task and the probing task \cite{belinkovprobing}. A high degree of probe accuracy indicates that the model's features possess sufficient information to accomplish the probing task.

To evaluate whether DPR training improved BERT's discriminative features, linear probing was employed on both pre-trained and DPR-trained BERT. A probe 

\[g_{lN}(f_{lq}, f_{ltp}, f_{lhn1}, f_{lhn2}, …) 
\]

was trained to classify which passage (between two, three, four, or five passages) is most relevant to a query. The Natural Questions dataset was used in this experiment, which provides labels for which passages are hard-negatives. The probe $g_{lN}$ at layer $l$ received BERT features for the query, true positive passage, and $N$ hard negative passages. $f_{lq}$ represents the features at layer $l$ for the query, $f_{ltp}$ represents the features for the true positive paragraph at layer $l$, and $f_{lhnN}$ represents the features for the Nth hard negative passage at the same layer. A distinct probe $g_{lN}$ was trained for each layer of BERT to examine how performance fluctuates across layers and with different numbers of hard-negative passages, thereby assessing how performance is impacted as the task's difficulty increases.

\begin{figure*}[t]
\centering
\includegraphics[width=\textwidth]{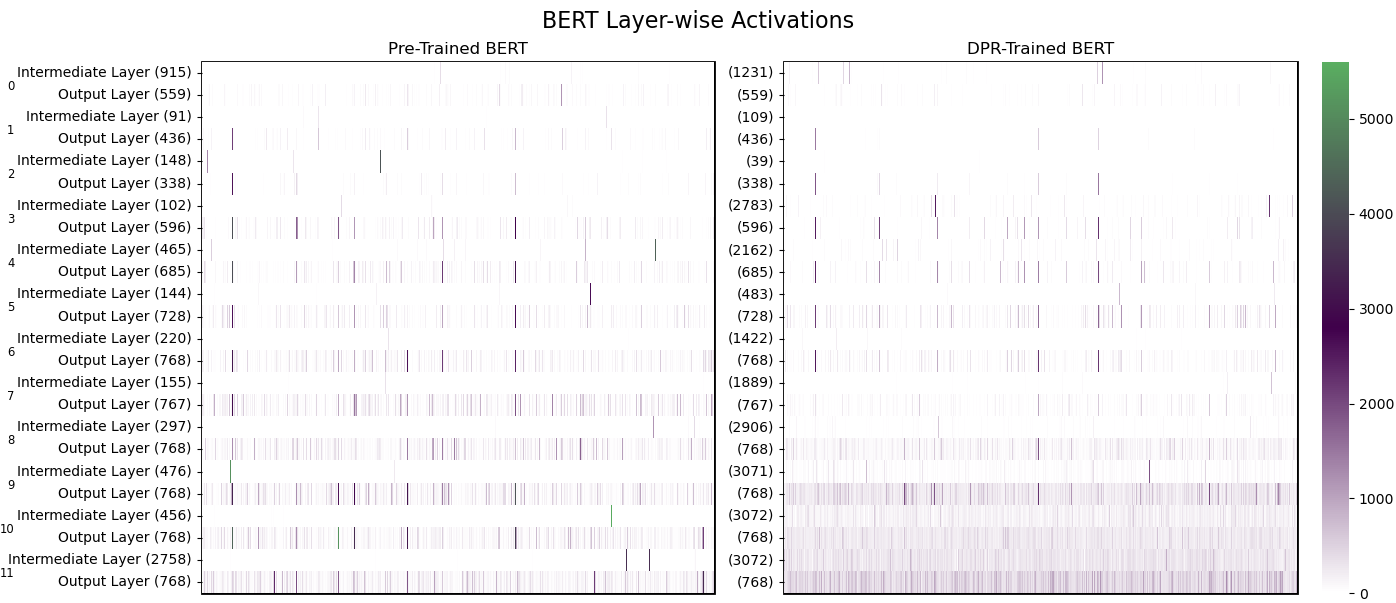}
\caption{Layerwise activations for pre-trained and DPR-trained BERT. The parenthetical numbers indicate the number of neurons in the layer that are above the attribution threshold for any number of examples.}
\label{fig:4layer_activation}
\end{figure*}

The difference between a true positive passage and a hard-negative passage is usually the presence of 1-2 key distinct facts in the passage. The ability to discriminate between 2-5 of these passages \textbf{indicates that the model likely has enough knowledge to know which facts are relevant to the query. This awareness is likely driven by the model's knowledge of the subject} (as discussed in later sections of this paper). Rather than testing overall retrieval ability, this experiment aims to find how aware/knowledgeable pre-trained BERT's features are compared to DPR-trained BERT when the difference of knowing or not knowing 1-2 facts can impact the final matching prediction.

Table \ref{tab:probing} shows the result of this experiment. The performance disparity between probes for pre-trained BERT and DPR-trained BERT is relatively minor in the two-passage scenario ($1.8\%$) and interestingly, it is the pre-trained BERT that exhibits a slight advantage. As the number of passages increases, the performance gap widens to approximately $6\%$, and overall probe efficacy declines. \textbf{These findings suggest that the inherent capability to discern relevant from irrelevant passages, when they are directly presented to the model, are likely already present in pre-trained BERT, and DPR-style training does not substantially enhance these discriminative features.} 

\section{Knowledge Decentralization in DPR-Trained Models}
\label{sec:knowledge_decentralization}
%While probing showed that the discriminative abilities before and after fine-tuning were similar, DPR-style training alters BERT's internal representation. 
The next perspective examined neuron activation patterns for the pre-trained and DPR-trained models. The knowledge attribution method from \cite{knowledgeneuron} was employed which was inspired by the pruning literature \cite{hao2021self,sundararajan2017axiomatic}. Our analysis targeted linear layers, as this is where the model stores knowledge according to prior research \cite{ffl_kvm}. 

To calculate an individual neuron’s contribution to the output, we varied its weight $w_i^{(l)}$ from $0$ to its original value. This can be calculated by:

\[\textrm{Attr}^{(l)}(w_i) = w_i^{(l)} \int_{\alpha=0}^{1} \frac{\partial P_x(\alpha w_i^{(l)})}{\partial w_i^{(l)}} \mathrm{d}\alpha \]

The Riemann approximation was used due to the intractability of calculating a continuous integral. Following \cite{knowledgeneuron}, a threshold of $0.1*\max(\textrm{Attr})$ was applied to identify a coarse set of knowledge neurons\footnote{Appendix \ref{sec:appendix2} demonstrates that our observations are consistent across a spectrum of thresholds.}. In contrast to \cite{knowledgeneuron}, the coarse set of knowledge neurons was not refined to a fine set of knowledge neurons, as our interest is in the broader activation patterns. When the model is processing inputs, both "true-positive" and "false-positive" knowledge neurons are activated indiscriminately with the activation strength corresponding to their attribution scores. The primary interest lies in how DPR training influences these activation patterns, rather than the role of specific neurons.

\begin{table*}[htbp]
{
\small
\centering
\begin{tabular}{m{2cm}|m{0.75cm}m{0.75cm}|m{0.75cm}m{0.75cm}|m{3cm}|m{3cm}}
\toprule
\multicolumn{1}{c}{} & \multicolumn{2}{c}{Answer in Top-1?} & \multicolumn{2}{c}{\# Strongly Activated Neurons} & \multicolumn{2}{c}{Title of Top-5 Retrieval} \\
 Query & Pre-trained BERT & DPR-BERT &  Pre-trained BERT & DPR-BERT & Pre-trained BERT & DPR BERT  \\
\midrule
where is the most distortion on a robinson projection & \xmark & \xmark & 220 & 1323 & Circle of latitude, Scale-invariant feature transform, Line moir{\'e}, Theil–Sen estimator, Pole splitting & Robinson projection, Robinson projection, Arthur H. Robinson, Robinson projection, Arthur H. Robinson \\
\hline
who is the chief legal advisor to the government & \xmark & \xmark & 65 & 831 & Jimly Asshiddiqie, Judicial system of Iran, Comptroller General of the State Administration, Jimly Asshiddiqie, Law of Kosovo & Attorney General of India, Attorney general, Attorney General of India, K. K. Venugopal, Attorney General of India \\
\hline
what type of government does kenya have 2018 & \cmark & \xmark & 74 & 287 & Government of Kenya, Abundant Nigeria Renewal Party, 2007–08 Kenyan crisis, Independent Electoral and Boundaries Commission, Kingdom of Kongo & Government of Kenya, Politics of Kenya, Government of Kenya, Government of Kenya, Government of Kenya \\
\hline
are pure metals made of atoms or ions & \cmark & \xmark & 69 & 1268 & Alloy, Common attributes, Metal, Resonance ionization, Alloy & Properties of metals, metalloids and nonmetals, Properties of metals, metalloids and nonmetals, Solid, Metal, Metal \\
\hline
who is the bad guy in lord of the rings & \xmark & \cmark & 100 & 533 & Millennium Earl, The Sword of Shannara, Eye of Ra, The Enchanted Apples of Oz, Ys I \& II & Saruman, Saruman, Sauron, Morgoth, Legolas \\
\hline
when were manatees put on the endangered list & \xmark & \cmark & 42 & 1522 & Ivory trade, Namib Desert Horse, Endangered Species Act of 1973, Iriomote cat, Bile bear & Manatee conservation, Endangered Species Act of 1973, Endangered Species Act of 1973, Manatee conservation, Endangered Species Act of 1973 \\
\hline
when did wesley leave last of the summer wine & \cmark & \cmark & 38  & 1024 & Naif (band), Aiden, Queensr{\"y}che, Josef Brown, Matthew Stocke & Gordon Wharmby, Gordon Wharmby, Brian Wilde, Cory Monteith, Last of the Summer Wine \\
\hline
when did mozart compose his first piece of music & \cmark & \cmark & 74 & 364 & Wolfgang Amadeus Mozart, Der Messias, Life of Franz Liszt, Die Entf{\"u}hrung aus dem Serail, Quattro versioni originali della \"Ritirata notturna di Madrid\" & Wolfgang Amadeus Mozart, Wolfgang Amadeus Mozart, Leopold Mozart, Wolfgang Amadeus Mozart, Wolfgang Amadeus Mozart \\
\bottomrule
\end{tabular}
}
\caption{This table presents example queries alongside the corresponding model retrievals and the count of strongly activated neurons for both pre-trained and DPR-trained BERT. Notably, DPR training consistently increases the number of strongly activated neurons. Additionally, the retrievals, even when DPR does not retrieve the correct passage in the top-1 retrieval, are much more focused and targeted to the asked query. In contrast, pre-trained BERT's retrievals are much more varied and sporadic. This is likely because in pre-trained BERT each neuron that is activated is responsible for more information and the model has no fine-grained path to follow for specific information like it does after DPR training.}
\label{tab:dpr_retrieval_examples}
\end{table*}

Figure \ref{fig:4layer_activation} illustrates the impact of DPR training on BERT’s neuron activations, charting the attribution score of every neuron across both the intermediate and output linear layers within each transformer block for the query model\footnote{Appendix \ref{sec:appendix} shows that the observations made in this section also hold for the context model.}. DPR-trained BERT has more activated neurons in the intermediate layer of each block. The output layer, on the other hand, maintains a consistent number of activations at each transformer block compared to pre-trained BERT, and in the earlier layers DPR-trained BERT activates fewer neurons in the output layers. Previous studies have conceptualized intermediate layers as "keys" and the output layer as the "value" \cite{ffl_kvm}. This suggests that \textbf{DPR training expands the set of "keys" available to access a given volume of semantic knowledge while decreasing the accessible volume of syntactic knowledge, embodying a decentralization strategy for semantic knowledge.} Rather than relying on a single, highly precise key to unlock some knowledge, DPR allows for the use of multiple, somewhat less precise keys. This underscores DPR training's primary goal: \textbf{to modify the model's method of knowledge access without altering the stored knowledge itself. These multiple pathways enable morphologically distinct but semantically related text to trigger the same knowledge or collections of facts, thus making retrieval possible.}

Table \ref{tab:dpr_retrieval_examples} demonstrates the effects of DPR training through performing retrieval with various queries and the full corpus of 21M Wikipedia passages. Across all instances, DPR training increases the number of strongly activated neurons indicating the existence of more pathways in the network allowing for better access to the information needed to perform retrieval. When examining the titles of the retrieved passages, a marked difference is revealed between pre-trained BERT and DPR BERT. Pre-trained BERT's retrievals are often disparate, aligning with the query in some instances while seemingly unrelated in others. This inconsistency indicates that successful retrievals by pre-trained BERT may hinge on the activation pattern precisely aligning with the relevant article. On the other hand, DPR-BERT, consistently retrieves passages that are topically related to the query, even if they are not the exact best match, reflecting a better ability to hone in on pertinent information. \textbf{By having more neurons responsible for each query the model has more fine-grained control to find relevant passages, even if what is found is not the most relevant possible passage.} In the cases where it was not able to navigate to the exact correct passage it is possible that the knowledge needed to discern between the correct and incorrect passage is not in the model.

\section{Knowledge Editing Experiments}
\label{sec:model_editing}
If DPR rearranges knowledge in the pre-trained BERT model, are these facts discoverable in DPR-BERT? We employed model editing techniques to remove facts from pre-trained BERT to investigate this. 

Currently, models struggle to learn and update the knowledge within them. Model editing is a class of techniques that minimizes pre-trained models to add or remove knowledge without disturbing the overall network. The types of methods generally used for model editing can be broadly categorized as fine-tuning with regularization, direct model editing, meta-learning, and architectural \cite{modeleditingreview}. 

Owing to the emerging state of this subfield and the variability in results, we employed various model editing techniques. In selecting techniques, we prioritized locality-preserving methods that minimally altered the model. This was to facilitate clearer attributions of our findings to DPR training rather than to potential architectural modifications. This led to TransformerPatch, MalMen, and Mend being chosen to perform the model editing \cite{transformerpatch,malmen,mend}. TransformerPatch is an architecture-modifying technique that introduces a single parameter to the last layer for each fact added. The experiments involving TransformerPatch used a learning rate of $3$e-$5$. MalMen and Mend, on the other hand, are meta-learning techniques. They utilize hypernetworks, a type of network that learns to generate parameters or parameter shifts for another model \cite{hypernetworks}. These hypernetworks were trained using a learning rate of $1$e-$6$ and to only modify the last three layers of the BERT network.

\subsection{Knowledge Removal}
To select the facts for removal, we identified questions from the NQ dataset that both DPR-BERT and the probed pre-trained BERT correctly answered. For each of the identified questions, we removed one fact from BERT, synthesized by transforming each query-answer pair from the NQ dataset into a cohesive sentence with GPT-4. Furthermore, when necessitated by the editing methodology, GPT-4 was employed to generate 10-12 rephrasings of each sentence.

A total of 284 queries, which both DPR-BERT and the linear probes had accurately matched with their corresponding passages, were randomly selected. Given that the chosen model editing techniques did not provide a direct method to explicitly remove facts from BERT, we employed previously described techniques to "overwrite" BERT's knowledge. To generate factually incorrect statements, the factually correct query-answer pairs were provided to GPT-4, which was prompted to generate new factually incorrect sentences. These new sentences were used by the model editing techniques to overwrite existing knowledge.

Fact removal success was determined by using the linear probing task from Section \ref{sec:probing} on the edited pretrained model. If the probe was able to still pick out the correct passage after fact removal, that indicated that the fact or its associative network was not fully removed. If the network's features (and hence its internal knowledge) were no longer sensitive enough to complete the task, then the fact was likely removed.

\begin{table}[pt]
{
\small
\centering
\begin{tabularx}{\columnwidth}{lr}
\toprule
\textbf{Knowledge Editing Technique} & \textbf{DPR Removed} \\
\midrule
Transformer-Patch & 0.87 \\
MalMen & 0.81 \\
Mend & 1.00 \\
\bottomrule
\end{tabularx}
}
\caption{Outcomes of the knowledge removal experiments. Of the facts successfully removed from BERT, the "DPR Removed" column is the percentage of the facts that were also absent after DPR training.}
\label{tab:knowledge_removed}
\end{table}

Of the facts successfully removed from the pre-trained BERT model by the knowledge editing techniques, Table \ref{tab:knowledge_removed} shows that $81\%-100\%$ were also absent in DPR-BERT. When a fact is removed it and likely part of its associative network is fully removed. As DPR training primarily functions to decentralize knowledge in a network, if the fact does not exist in the network there is nothing to decentralize in the network. DPR training does not appear to add knowledge to the network, once a fact is removed, it cannot be recovered through DPR training.

% This limited success could stem from the complexity of fully erasing a fact, given that facts are interdependent, exist in multiple logical forms, and are supported by neighboring facts that might compensate for any inaccuracies introduced. This complexity, along with the fact that existing facts are being overwritten rather than new ones being introduced, may contribute to the higher incidence of off-target edits when performing fact removal. Notably, the overwritten facts appear to be more strongly set into BERT. $81\%-100\%$ of the facts that are overwritten were also incorrectly matched in DPR-BERT, as shown in Table \ref{tab:knowledge_removed}. This outcome suggests that once a fact and its interconnected network are overwritten, the ability to train a model to retrieve context that requires that fact becomes significantly compromised. It is unlikely that post-removal the fact remains in the network in a form that can be decentralized in a way that makes it retrievable. 

The knowledge removal experiment demonstrates that \textbf{DPR training primarily refines how pre-existing knowledge within BERT is rendered more "retrievable".} Thus, it appears that DPR training \textbf{does not alter the model's inherent knowledge base; instead, it modifies the representation and accessibility of this knowledge.}

\section{Related Works}
DPR addresses the challenge of matching a query with the most relevant passages from a knowledge base \cite{dpr}. This approach employs dual encoders—one encoder for the passages and another for the query—and utilizes a distance metric, such as the inner product, to identify the passages closest to the query. Inspired by Siamese networks \cite{siamese}, DPR represents the first fully neural architecture to outperform the BM25 algorithm \cite{bm25}. Since then, there have been quite a few improvements in how to train DPR-style models. Methods like RocketQA improve DPR by employing cross-batch negatives and training the network on more difficult hard negatives \cite{RocketQA}. Dragon focuses on novel data augmentation and supervision strategies \cite{Dragon}. Contriever also employs a greater number of hard-negatives and data-augmentation methods in addition to pre-training the model on the inverse cloze task \cite{contriever}.  MVR generates multiple views for each document to allow for multiple diverse representations of each of them \cite{MVR}. ColBERT employs token embeddings for more fine-grained matching \cite{Colbert}. REALM leverages feedback from the reader component to jointly train the retriever with the reader \cite{Realm}. Other methods distill knowledge from the reader to the retriever \cite{izacard2020distilling,distillation2023}. Additionally, efforts in query augmentation or generation aim to better synchronize the query with the document encoder \cite{queryrewriting1,query2doc,iterretgen,hyde}. Despite these different enhancements, each method builds upon the DPR framework discussed in this paper.

Distinctly, RetroMAE and CoT-MAE pre-train a model using a masked auto-encoder strategy, which they show enhances downstream retrieval performance \cite{retromae,cotmae,retromaev2,cotmaev2}. Following this pre-training phase, both methods subsequently adopt DPR fine-tuning to further refine their models for improved task performance.

Only a few studies have delved into analyzing DPR models. One such study took a holistic look at RAG to see where the pipeline made errors \cite{blamegame}. The study found that a similarity-based search during retrieval biased the result in favor of passages similar to the query, even when more relevant but dissimilar passages were available. Another study employed probing techniques to analyze ranking models \cite{abnirml}. The authors adopted a probing method akin to ours, categorizing passages by specific properties for analysis, in contrast to our approach of random selection among hard negatives. This study explored how query and document characteristics affect ranking outcomes. Another study analyzed the embeddings produced by retrieval models in the vocabulary space \cite{tokenabout}. To do this, they used pre-trained BERT’s MLM head on the DPR-trained embeddings' [CLS] token. It was found that DPR implicitly learns the importance of lexical overlap between the query and passage. DPR training causes BERT to retrieve passages that share more tokens with the query as compared to pre-trained BERT. This ties in with our finding where the number of output layer activations in the early part of the model post-DPR training decreased. This may function as a sort of syntactic filter, where many keys can access fewer, but more pertinent, lexical features. However, this filtering can also induce what the author's term “token amnesia”. This condition occurs when an encoder fails to correctly retrieve relevant passages because it does not properly encode the relevant token, usually related to a named entity. Unlike previous research, our study adopts a holistic approach, examining model knowledge, activation patterns, and capabilities across different model stages. This analysis approach integrates and makes sense of the different insights from prior works.

\section{Conclusion}
This paper analyzes the effect of DPR fine-tuning using different analysis methods. Through these analyses, we find that: 
\begin{enumerate}
    \item DPR training alters how knowledge is stored in the model (Section \ref{sec:knowledge_decentralization}).
    \item DPR likely does not change how much knowledge is contained in the model (Section \ref{sec:probing}).
    \item DPR's ability to retrieve knowledge is an extension of the knowledge contained in the model (Section \ref{sec:model_editing})
\end{enumerate}
DPR training refines and restructures how the model stores information in the network transitioning from a centralized pattern of storage to a decentralized pattern. This decentralization makes it so that each fact/memory has a lot more pathways to get triggered, which in turn allows for more potential inputs to trigger the same set of memories. By engaging more neurons, more robustly for each fact, the model diminishes the uniform reliance on a specific neuron for a specific fact. This allows the model to embed different semantic forms of a query into a vector that will still map to a similar set of relevant passages, instead of requiring a specific ordering of words to trigger a specific neuron in order to get the relevant passage.

Additionally, we find that BERT does not acquire new knowledge through DPR fine-tuning. Knowledge can be removed before DPR fine-tuning to enhance the model, but the training process \textbf{does not} seem to change the knowledge content of a model. Facts that are removed stay removed and are unreachable. Being able to discriminate between two pieces of text, like in the linear probing experiment of Section \ref{sec:probing}, in a classification task is about using the knowledge contained in the network to tell two pieces of text apart. The knowledge to understand the one to two fact difference in pairs (or more) of text is apparent in the features of both pre-trained and DPR-trained BERT. This piece of evidence indicates that DPR's aim is not to learn new facts, but to align disparate pieces of text in a common embedding space and restructure how knowledge is accessed to facilitate that. 

If DPR does not add facts, then we should be able to remove relevant facts that pre-trained BERT knows and not find them again in DPR-trained BERT. Section \ref{sec:model_editing} indicates that this statement is true, suggesting from another perspective, that DPR is dependent on the knowledge already contained in the model to produce satisfactory retrievals. If a fact is removed, it will likely not be added back during DPR training. Thus the existence or absence of necessary facts or webs of knowledge within a model can aid or hamper its ability to retrieve information.

In the most fundamental sense, DPR achieves its namesake function—it retrieves, locating and returning relevant context to the user given a query. Yet, as our evidence suggests, DPR models appear constrained to retrieving information based on the knowledge that preexists within their parameters, either innately or through augmentation. This operational boundary delineates a significant caveat: facts must already be encoded within the model for useful context to be accessible by retrieval. Absent these facts or their associative networks, retrieval seems to falter. Thus, if retrieval is understood as the capacity to recall or recognize knowledge already familiar to the model or on the periphery of what is familiar, then indeed, DPR models fulfill this criterion. However, if we extend our definition of retrieval to also encompass the ability to navigate and elucidate concepts previously unknown or unencountered by the model—a capacity akin to how humans research and retrieve information—our findings imply that the current DPR model fall short of this mark.

Our findings suggest several areas of focus for future work including (1) 
accelerate knowledge representation decentralization with new unsupervised training methods (2) %Given a network that already stores knowledge in a decentralized representation, 
develop new methods to inject facts in a decentralized manner into the network (3) optimize retrieval methods that operate with uncertainty, and (4) map the model's internal knowledge directly to the set of best documents to retrieve. 

Current work in optimizing the inverse cloze pre-training task and various data augmentation methods such as \cite{Dragon} begin to address (1) by increasing the amount of knowledge that the model is exposed to during fine-tuning and thus the amount that can be decentralized. With the knowledge of the purpose of DPR-training more targeted methods can be developed. (3) requires more detailed analysis to determine how a model processes a query when it is missing key knowledge needed for retrieval. Being aware of when a model is uncertain in its retrieval is crucial. The analysis should reveal methods to more robustly and gracefully handle increased levels of uncertainty. One direction to better leverage a model’s knowledge as suggested in (4) is shown in \cite{dsi, dsiscale, nci, seal, llmautoregressive}.

\section{Limitations}
This paper presents a detailed analysis of the DPR formula, specifically focusing on the original DPR training formula utilizing a BERT backbone. We anticipate that our findings will exhibit a degree of generalizability across various DPR implementations, given the underlying commonalities of the core training approach. It is important to recognize that modifications—such as improving hard negatives, different data augmentation techniques, different transformer-based backbones, or leveraging multiple views/vectors from models—while serving to refine and enhance the DPR framework, build upon and amplify the mechanisms of the DPR method. These enhancements, though significant in optimizing performance, are not expected to fundamentally change this analysis. However, it is still a limitation of this paper that we did not repeat our analysis on more DPR-based methods and datasets. 

\section{Ethics Statements}

This work presents an analysis of DPR-style training. Improving DPR-style training would improve RAG pipelines, increasing the factuality of LLMs and decreasing the rate which they hallucinate.

\bibliography{custom}
\appendix

\section{Appendix}

\subsection{Context Model Activations}
\label{sec:appendix}

\begin{figure*}[t]
\centering
\includegraphics[width=\textwidth]{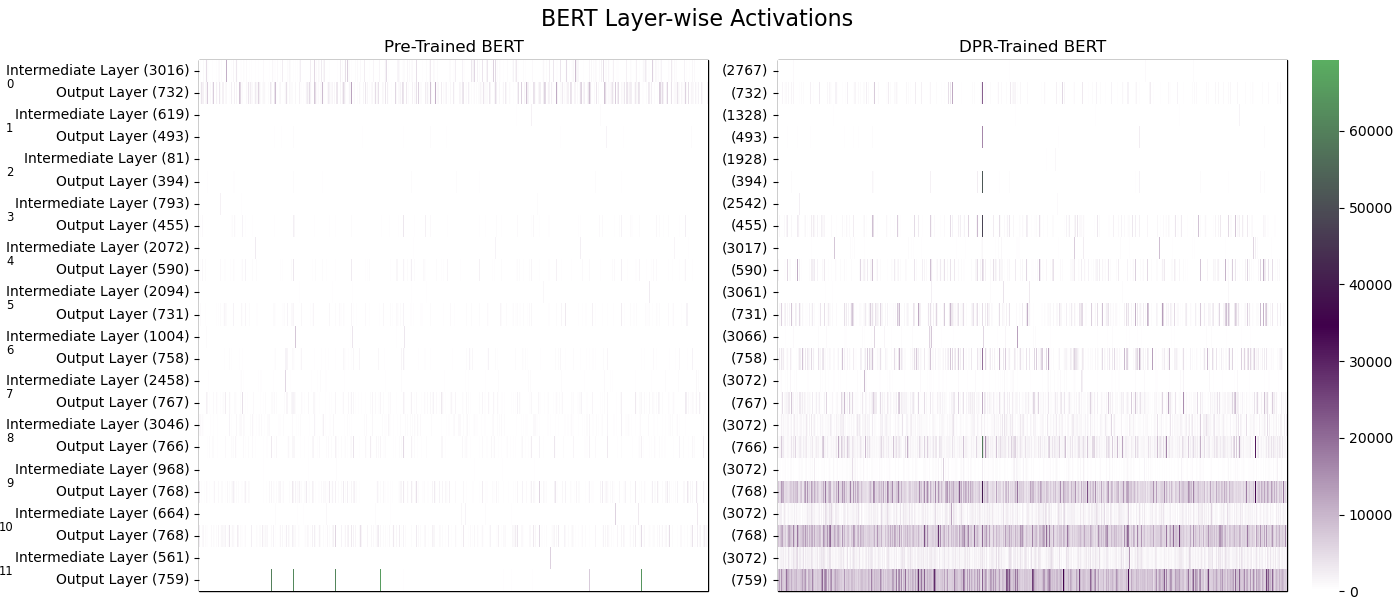}
\caption{Layerwise activations for pre-trained and DPR-trained BERT - context model. The parenthetical numbers indicate the number of neurons in the layer that are above the attribution threshold for any number of examples.}
\label{fig:5layer_activation_context_model}
\end{figure*}

Figure \ref{fig:5layer_activation_context_model} depicts the activation patterns observed in the context model, mirroring the trends outlined in Section \ref{sec:knowledge_decentralization}. The only exception occurs in the first intermediate layer of the pre-trained BERT model, where a larger number of neurons are activated as compared to DPR-trained BERT.

\subsection{Model Activations at different thresholds}
\label{sec:appendix2}

\begin{figure*}[t]
\centering
\includegraphics[width=\textwidth]{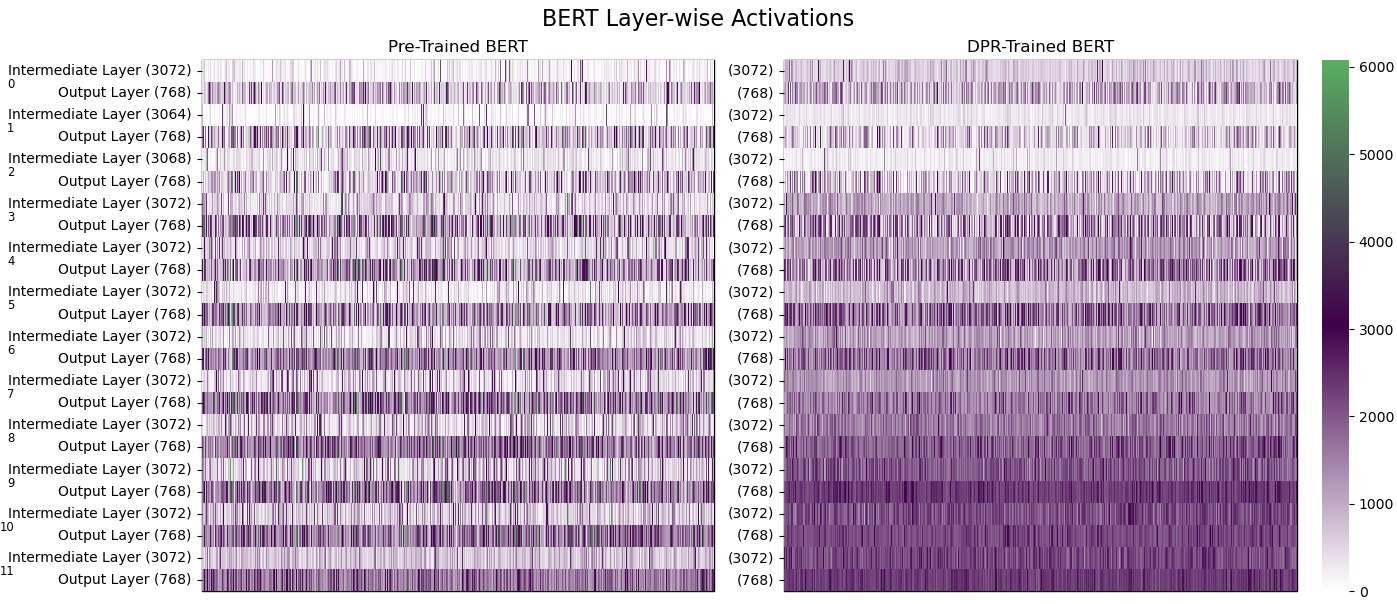}
\caption{Layerwise activations for pre-trained and DPR-trained BERT with a threshold of 0.005. The parenthetical numbers indicate the number of neurons in the layer that are above the attribution threshold for any number of examples.}
\label{fig:6layer_activation_0.005}
\end{figure*}

\begin{figure*}[t]
\centering
\includegraphics[width=\textwidth]{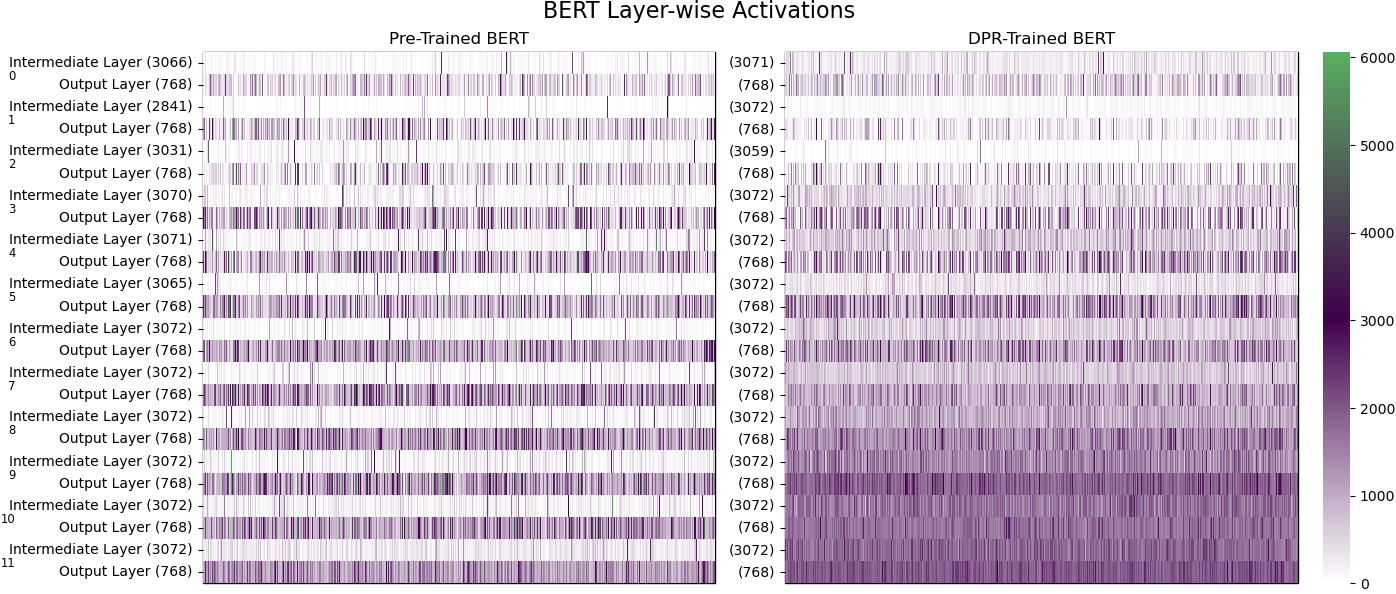}
\caption{Layerwise activations for pre-trained and DPR-trained BERT with a threshold of 0.01. The parenthetical numbers indicate the number of neurons in the layer that are above the attribution threshold for any number of examples.}
\label{fig:7layer_activation_0.01}
\end{figure*}

\begin{figure*}[t]
\centering
\includegraphics[width=\textwidth]{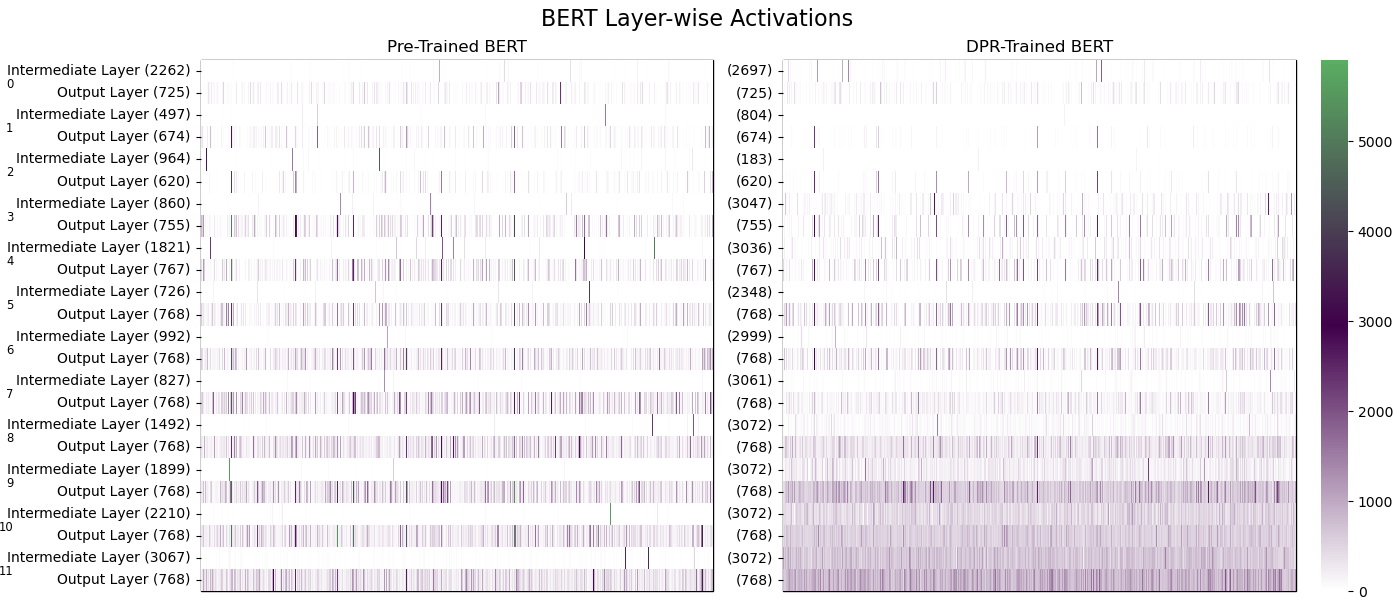}
\caption{Layerwise activations for pre-trained and DPR-trained BERT with a threshold of 0.05. The parenthetical numbers indicate the number of neurons in the layer that are above the attribution threshold for any number of examples.}
\label{fig:8layer_activation_0.05}
\end{figure*}

\begin{figure*}[t]
\centering
\includegraphics[width=\textwidth]{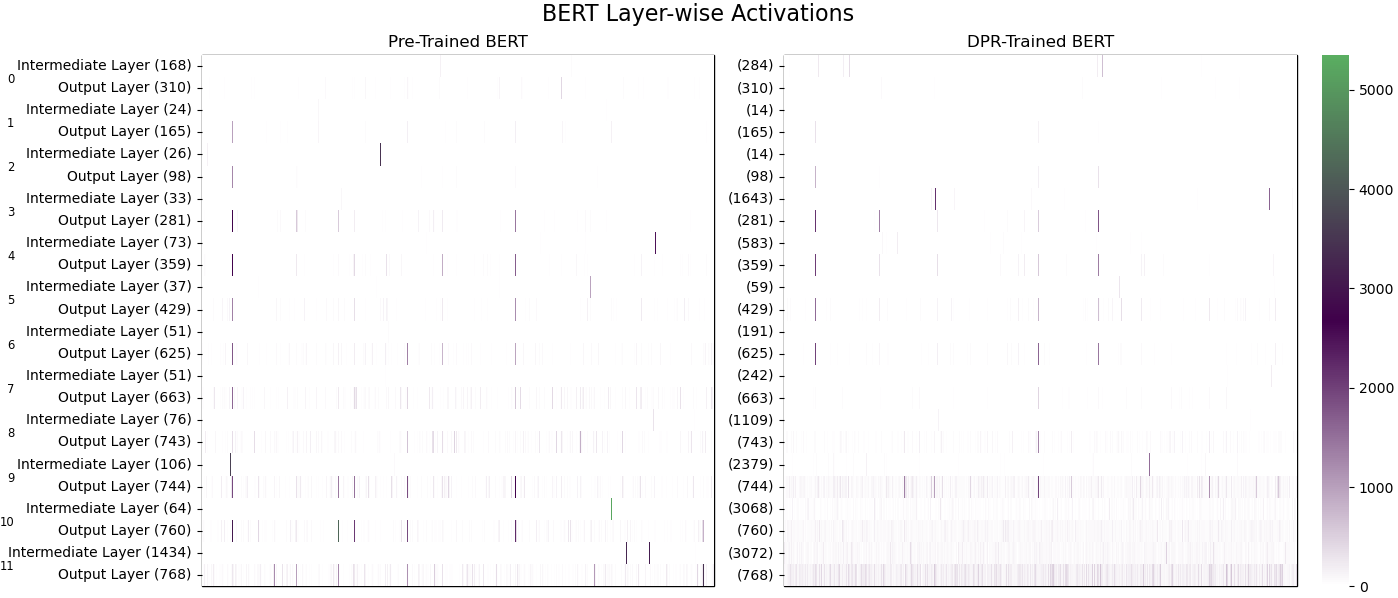}
\caption{Layerwise activations for pre-trained and DPR-trained BERT with a threshold of 0.2. The parenthetical numbers indicate the number of neurons in the layer that are above the attribution threshold for any number of examples.}
\label{fig:9layer_activation_0.2}
\end{figure*}

\begin{figure*}[t]
\centering
\includegraphics[width=\textwidth]{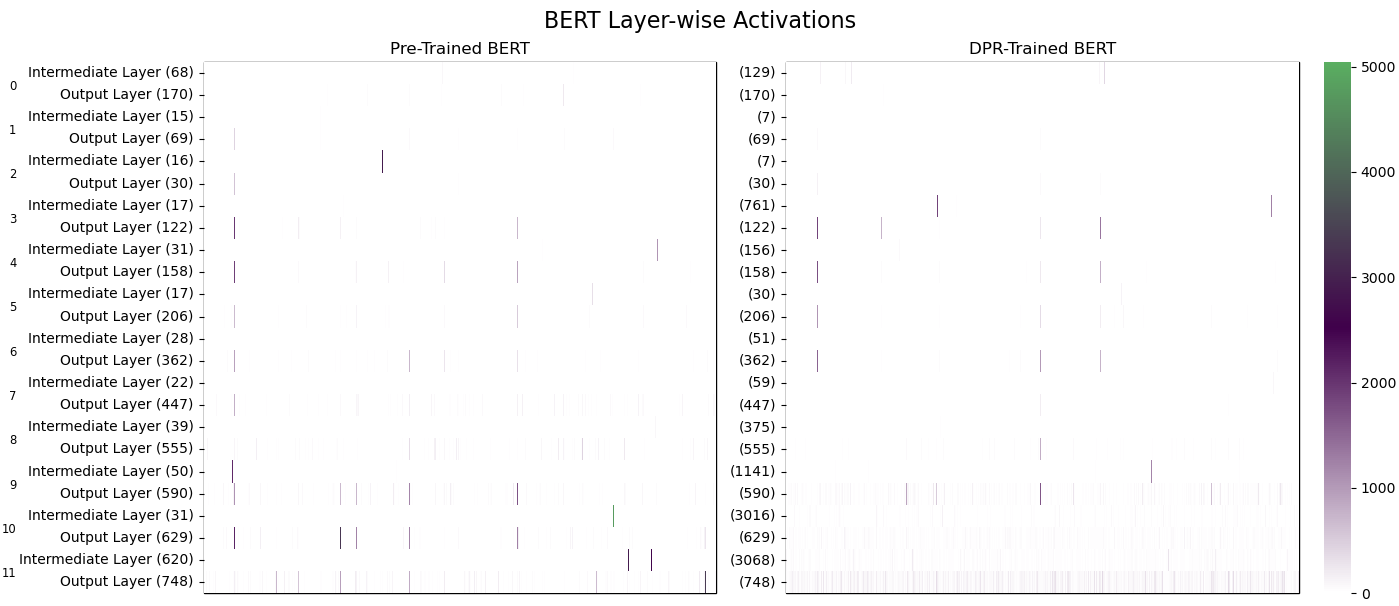}
\caption{Layerwise activations for pre-trained and DPR-trained BERT with a threshold of 0.3. The parenthetical numbers indicate the number of neurons in the layer that are above the attribution threshold for any number of examples.}
\label{fig:10layer_activation_0.3}
\end{figure*}

Figures \ref{fig:6layer_activation_0.005}, \ref{fig:7layer_activation_0.01}, \ref{fig:8layer_activation_0.05}, \ref{fig:9layer_activation_0.2}, and \ref{fig:10layer_activation_0.3} illustrate neuron activation patterns across varying activation thresholds set at $0.005*\max(Attr)$, $0.01*\max(Attr)$, $0.05*\max(Attr)$, $0.2*\max(Attr)$, and $0.3*\max(Attr)$, respectively. As the threshold increases from 0.005 to 0.3, the visualization narrows down to neurons with stronger activations. This observation reinforces the findings discussed in Section  \ref{sec:knowledge_decentralization}: pre-trained BERT shows a trend of fewer but more consistently activated neurons, in contrast to DPR-trained BERT, which exhibits a broader array of neurons activated less frequently.

\end{document}